\documentclass[conference]{IEEEtran}
\IEEEoverridecommandlockouts
\usepackage{cite}
\usepackage{booktabs}
\usepackage{amsmath,amssymb,amsfonts}
\usepackage{algorithmic}
\usepackage{graphicx}
\usepackage{xcolor, soul} 
\usepackage{textcomp, balance}
\usepackage{xcolor}
\def\BibTeX{{\rm B\kern-.05em{\sc i\kern-.025em b}\kern-.08em
    T\kern-.1667em\lower.7ex\hbox{E}\kern-.125emX}}

    \sethlcolor{yellow}
\begin{document}
\title{Metaverse: Requirements, Architecture, Standards, Status, Challenges, and Perspectives\thanks{ This work was partly supported by the DoD CoE-AIML at Howard University under Contract W911NF-20-2-0277 with the U.S. Army Research Laboratory, Microsoft Corp.  Research Gift Funds and Meta/Facebook Research Gift Funds.\\
\textbf{\color{red} This paper is accepted for publication by IEEE Internet of Things Magazine's Special Issue.}}}
\author{\IEEEauthorblockN{Danda B. Rawat and Hassan El alami}
\IEEEauthorblockA{Howard University, Washington, DC, USA \\ 
\{danda.rawat, hassan.elalami\}@howard.edu}}
\maketitle

\begin{abstract}
The Metaverse is driving the next wave of innovation for new opportunities by replacing the digital world (Internet) with the virtual world through a single, shared, immersive, persistent 3D virtual space. In this paper, we present requirements, architecture, standards, challenges, and solutions for Metaverse. Specifically, we provide Metaverse architecture and requirements, and different standards for Metaverse which  serve as the basis for the development and deployment. Moreover, we present recent status, challenges such as integration of AI and Metaverse, security and privacy in Metaverse, etc., and perspectives and solutions.
\end{abstract}


\section{Introduction}
Metaverse is anticipated to be one of the most significant technological advances of the coming years. Essentially, the Metaverse consists of computer-generated worlds with independent economies and consistent value systems. The Metaverse provides users with immersive cyber-virtual experiences in physical worlds by connecting a single, universal Internet to an integrated 3D virtual and physical world network. Particularly, two popular applications, augmented reality (AR) and virtual reality (VR), are being developed to make the Metaverse an immersive digital experience and a social network for its users. Additionally, the Metaverse is recognized as the next-generation Internet (3D Internet), in which users have the opportunity to live as digital natives and have the opportunity to experience an alternative lifestyle through a virtual environment~\cite{R5}.  Moreover, various realistic scenarios for using the Metaverse are available in various fields, such as education, healthcare, tourism, etc. As an example, the use of the Metaverse in healthcare has the potential to significantly improve patient outcomes by enabling completely new methods of providing cost-effective treatments. In addition to allowing remote monitoring and telemedicine for geographically dispersed patients, the Metaverse can also be an excellent tool for healthcare professionals. Additionally, healthcare KPIs and metrics, such as treatment costs, treatment error rate, and patient satisfaction can be improved through the use of the Metaverse in healthcare.

The Metaverse integrates a variety of emerging technologies, including AR/MR/VR/XR, Digital Twins, 5G/6G, Internet of Things (IoT), Artificial Intelligence (AI), and security and privacy to ensure its success. Moreover, COVID-19 is one of the main driving forces behind the excitement surrounding the Metaverse. It has led to fundamental changes in how people work, entertain themselves, and socialize today~\cite{3R}. Furthermore, the Metaverse has been positioned as an essential component of the future as users become more accustomed to performing conventional tasks remotely. As part of web 3.0, the Metaverse is recognized as a paradigm that is evolving~\cite{IT}. In accordance with~\cite{R1}, as illustrated in Fig.~\ref{Fig:Char}, avatars are able to seamlessly traverse a variety of virtual worlds, including sub-Metaverses, in order to experience a digital environment, as well as participate in virtual economic activities through physical infrastructures and a Metaverse incentive. Six characteristics of the Metaverse can be summarized as follows:
\begin{figure}[!b]
 \includegraphics[width=\linewidth]{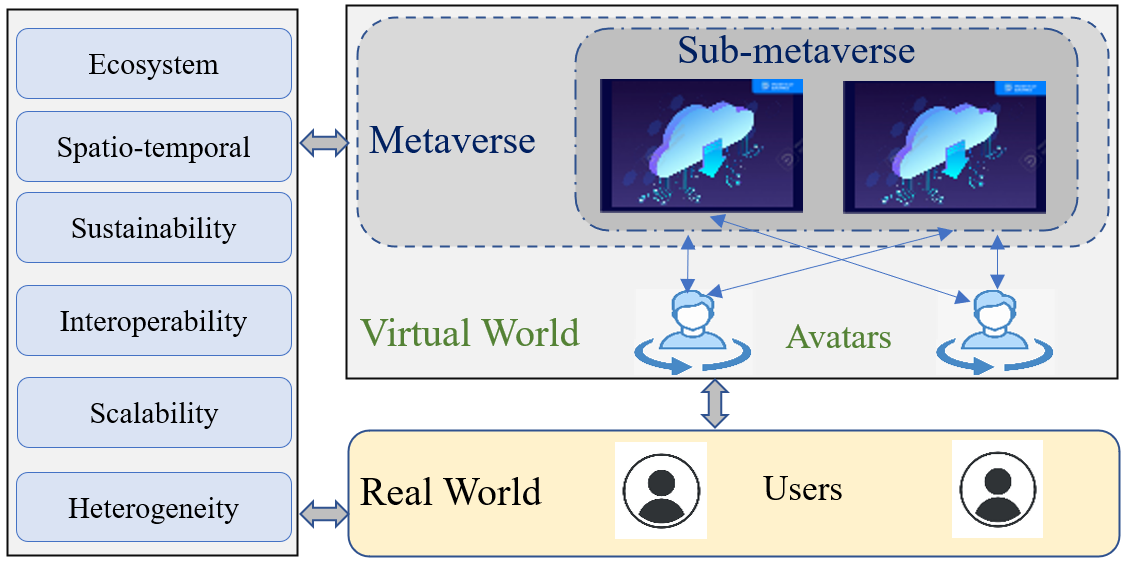}
 \caption{Characteristics of the Metaverse.}
 \label{Fig:Char}
\end{figure}
\\
\textbf{Immersiveness:} It refers to a computer-generated virtual environment that is sufficiently realistic for users to feel psychologically and emotionally involved. Through sensory perception, such as sight, sound, touch, temperature, and pressure, as well as through expressions, such as gestures and signs, it is possible to understand it.\\
\textbf{Spatio-temporal:} There are limits to the real world due to the finiteness of space and the irreversibility of time. The spatiotemporal refers to the break of space and time limitations in the Metaverse, a virtual space-time continuum parallel to the real one.\\
\textbf{ Sustainability:} A high degree of independence and a closed economic cycle indicate that the Metaverse maintains a consistent value system and a closed economic loop.\\
\textbf{Interoperability:} It implies that users can seamlessly move between virtual worlds (i.e., sub-Metaverses) without interrupting the immersive experience. Moreover, digital assets used for rendering or reconstructing virtual worlds can be interchanged between different platforms.\\
\textbf{Scalability:} It refers to the Metaverse's ability to remain efficient with the number of concurrent users/avatars, the level of scene complexity, scope, and range of interactions between users/avatars.\\
\textbf{Heterogeneity:} In the Metaverse, heterogeneous virtual spaces exist such as with distinct implementations, heterogeneous physical devices exist such as with different interfaces, heterogeneous data types exist, heterogeneous communications modes exist, and human psychology is diverse.

In this paper, we present a holistic and comprehensive overview of the Metaverse, including its requirements, architecture, and standards, as well as its challenges, open problems, and perspectives.

\section{Metaverse Architecture}
As the Metaverse is currently in its early stages of development, academia and industry do not have a consistent definition of its architecture. Based on~\cite{R1},~\cite{35}, we propose that the Metaverse architecture spans the real and virtual worlds. Accordingly, the proposed architecture can be used to encompass, as shown in Fig.~\ref{Fig:Mv}, the following components of a Metaverse:

\begin{figure*}[!t]
    \centering
        \includegraphics[width=.65\textwidth]{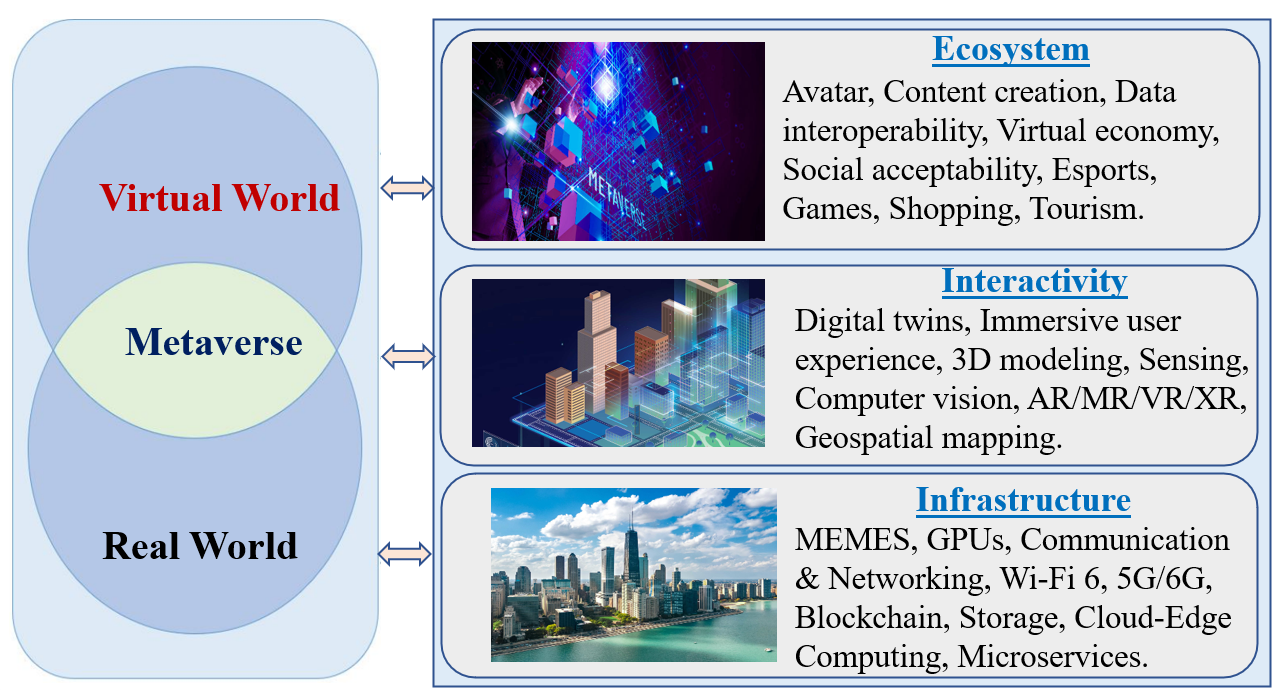}
    \caption{Typical architecture of the Metaverse.}
    \label{Fig:Mv}
    \vskip-0.45cm
\end{figure*}

\begin{itemize}
\item \textbf{Infrastructure:} The infrastructure  includes the fundamental technologies that support the operation of the Metaverse, such as MEMS, GPUs, communication and networking (e.g., Wi-Fi 6, 5G/6G), blockchain,  cloud-edge computing, storage, and Microservices.
\item \textbf{Interactivity:} It serves as an interface between the virtual and real worlds, allowing the real world to be connected to the Metaverse and the Metaverse to be realized. It includes digital twins, immersive user experiences, 3D modeling, sensing, computer vision, Geo-spatial mapping, and AR/MR/VR/XR technologies. 
 \item \textbf{Ecosystem:} It facilitates the running of the entire Metaverse, including avatars, content creation, data interoperability, virtual economy, social acceptance, e-sports, gaming, shopping, tourism, and others.

\end{itemize}

 \section{Metaverse Standards}
Industry and standards communities are developing standards to make Metaverse concepts a reality. So, as shown in Table~\ref{tab:standard}, the existing Metaverse standards are primarily concerned with seamless connectivity between the real world and the virtual world.

\begin{itemize}
\item\textbf{ISO/IEC 23005:} It aims to standardize the interfaces between the physical world and the virtual world, as well as between virtual worlds, to achieve interoperability, simultaneous reaction, and seamless information exchange~\cite{37}. ISO/IEC 23005 is applicable to a wide range of Metaverse business services, where audiovisual information, rendered sensory effects, and virtual object characteristics can be integrated into interactions between real and virtual worlds.
\item\textbf{IEEE 2888:} IEEE 2888 defines standardized interfaces for synchronizing cyberspace and the physical world~\cite{38}. IEEE 2888 standards define information formats and application program interfaces (APIs) for controlling actuators and acquiring sensory information, allowing virtual and real worlds to interact.
\item\textbf{IEEE P1589:} IEEE 1589 is proposed for the augmented reality learning experience, which describes how activities, the learning context, specific environments, and potentially other elements of AR-enhanced learning activities will be represented in a standardized interchange format, along with data specifications~\cite{39}. By leveraging sensors and computer vision, it makes it easier for users to create learning experiences that integrate real-world interactions with web applications.
\item\textbf{IEEE P2048:} As the Metaverse industry is still in its infancy, there has been much hype, confusion, and misunderstanding about it. A lack of consensus on basic terminology, definitions, and taxonomies can mislead early adopters and create unnecessary barriers to progress. It is intended that IEEE P2048~\cite{40} will define the terminology, categories, and levels of a Metaverse, as well as facilitate the sustainable development of Metaverse-related activities and promote the healthy growth of the market for Metaverse.
\item\textbf{IEEE P7016:} The aims of this standard are to provide a high-level overview of the techno-social aspects of Metaverse systems and to specify an ethical assessment methodology for use in their design and operation~\cite{41}. It provides guidelines to Metaverse developers on how to prioritize ethically aligned system design. 
\end{itemize}
\begin{table}[!h]
   \caption{Summary of Metaverse Standards}
   \label{tab:standard}
   \centering
\begin{tabular}{|l|l|}
\hline
\bf{Standard}      & \bf{Description}                                                                                                                                                              \\ \hline
ISO/IEC 23005 & \begin{tabular}[c]{@{}l@{}}Provides guidelines for the design,  development, \\ and deployment of Metaverses.\end{tabular}                                                       \\ \hline
IEEE 2888     & \begin{tabular}[c]{@{}l@{}}Standardizes interfaces for the physical and virtual\\ worlds, and defines data formats and APIs.\end{tabular}                                      \\ \hline
IEEE P1589    & \begin{tabular}[c]{@{}l@{}}It is for AR learning experiences that facilitate \\ the creation of experience repositories and online\\ marketplaces.\end{tabular}                 \\ \hline
IEEE P2048    & \begin{tabular}[c]{@{}l@{}}Provides terminology, definitions, and taxonomies\\   in the Metaverse,  and ensures that the Metaverse\\  can develop sustainably.\end{tabular}       \\ \hline
IEEE 7016     & \begin{tabular}[c]{@{}l@{}}Presents an overview of the techno-social aspects\\ of Metaverses,  as well as a methodology for\\ evaluating their ethical viability.\end{tabular} \\ \hline
\end{tabular}
\end{table}

\section{Key Components and Requirements of Metaverse}
\begin{figure}[!b]
\vskip-0.45cm
 \includegraphics[width=\linewidth]{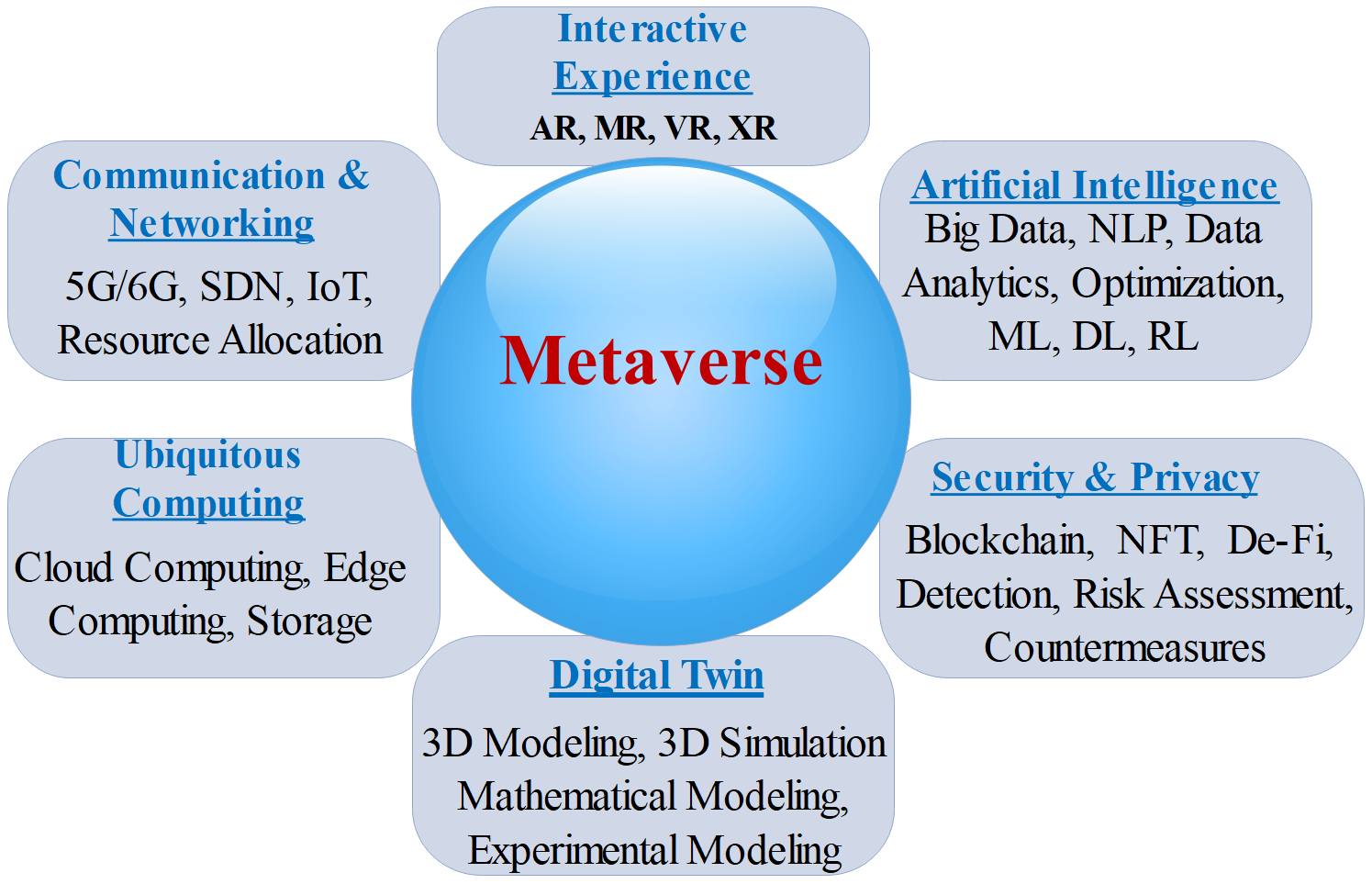}
 \caption{Illustration of key requirements and its role in the Metaverse.}
 \label{Fig:Req}
\end{figure}

A variety of new technologies are required in order to construct the Metaverse. This section presents the most significant key technological requirements which have been discussed in the selected literature. As shown in Fig.~\ref{Fig:Req}, there are the following requirements underlying the Metaverse.
\subsection{Communication and Networking}
A number of technologies are enabling real-time massive data communication between physical and virtual worlds, as well as between sub-Metaverses, such as 5G/6G, Software-Defined Networks (SDN), IoT, and Resource Allocation Frameworks. Besides 5G/6G, there are possibilities for real-time and ultra-reliable communications with enhanced mobility for massive Metaverse devices~\cite{42}. 
In the Metaverse, SDN facilitates flexible and scalable management of large-scale networks through the separation of control planes and data planes. The physical devices and resources of an SDN-based Metaverse are managed logically using a standardized interface like OpenFlow, allowing virtualized computation, storage, and bandwidth resources to be dynamically allocated based on the real-time needs of sub-Metaverses. The IoT sensors are extensions of the human senses in the Metaverse. The Metaverse requires enormous resources, especially computing resources for intensive data processing to support XR, massive storage resources, and massive networking resources to maintain ultra-high-speed and low-latency connections~\cite{45}. Therefore, resource allocation frameworks can manage and allocate different types of resources for Metaverse.
\subsection{Ubiquitous Computing}
By creating an environment where computing can be accessed anywhere at any time, ubiquitous computing aims to provide users with access to computing at any time~\cite{46}. Ubiquitous computing is enabled by the integration of ubiquitous smart objects into the physical environment, allowing for fluid adaptation to interactions between users and the physical environment. This technology allows users to interact freely with their avatars and experience immersive Metaverse services through ubiquitous smart objects and networks, rather than having to use specific equipment such as cell phones, laptops, and end devices. 

\subsection{Digital Twin}
 
It represents high fidelity and consciousness as a digital replica of the real world~\cite{48}. Analyzing real-time streams of data and physical models, as well as historical data, enables the reflection and optimization of virtual users. The digital twin learns and adapts to mirrored space on its own by feeding back data from real entities. Using 3D simulations and AI, it can produce highly accurate 3D models of expected objects in the Metaverse, which is beneficial to the creation and rendering of large-scale Metaverses. Mathematical and experimental modeling can also be used to design a virtual model of the physical system. Mathematical models generally make simplified assumptions (e.g., linear approximations for nonlinear functions). Experimental modeling can also be employed through a series of experiments. Providing bidirectional connectivity between the real and virtual worlds can enable predictive maintenance and accident tracking, reducing risks and improving efficiency.
\subsection{Interactive experience}
In light of the maturation of miniaturized sensors, embedded technologies, as well as XR technology, XR devices are expected to be the most significant terminal for entering the Metaverse. By integrating VR/AR/MR technologies, XR provides immersive experiences, augmented experiences, and real-time interaction with users, avatars, and environments through front-projected holographic displays, Human-Computer Interaction, and large-scale 3D modeling~\cite{50}. Specifically, VR offers the ability to immerse oneself in a virtual world, AR provides the ability to experience holograms, graphics, and video in real-time in the real world, and MR offers a transitional experience between these technologies.  Therefore, the interaction between a user and an  avatar will no longer be limited to mobile devices (e.g., cell phones and laptops), but will be available across a wide range of interactive devices. Furthermore, the combination of low-latency edge computing systems and AI-driven real-time rendering can eliminate negative experiences such as dizziness when wearing XR devices (e.g., headsets and helmets).
\subsection{Security and Privacy}
In light of the many potential applications of Metaverse in modern life, security and privacy represent significant key requirements. Metaverse may be subject to a variety of threats and attacks. As such, different cyber-defense techniques, such as detection, risk assessment, countermeasures, and privacy, must be taken into account to ensure the security and protection of data in the virtual and real worlds. Using blockchain technologies, the virtual economy can be built sustainably, and the value system in the Metaverse can be constructed. These technologies provide an open and decentralized solution. In the context of blockchain technology, a non-fungible token (NFT) technique is an irreplaceable and indivisible token, which can be used to help identify assets and trace ownership provenance on the chain. In addition, the concept of decentralized finance (De-Fi) is used in the Metaverse and is designed to provide secure, transparent, and highly complex financial services.

\subsection{Artificial Intelligence} 
By combining AI with other technological requirements, such as communication, ubiquitous computing, interaction, security \& privacy, the Metaverse is able to create secure, scalable, and realistic virtual worlds that are always available. Considering the Metaverse architecture (see Fig.~\ref{Fig:Mv}), it is undeniable that AI can enhance infrastructure reliability and improve its performance. For communication and networking technologies, a variety of advanced AI and optimization algorithms, ML, DL, and RL have been applied to address a variety of challenging problems, including energy efficiency, resource allocation, scalability, and security and privacy threats. By using IoT sensors/actuators and other human-machine interaction devices, simple human movements and complex actions can be analyzed and recognized based on ML/DL. As a result, users' physical movements in the real world are projected into the virtual world, allowing them to fully control their avatars in order to interact with other objects in the Metaverse. As well as interacting with many modalities used in the physical world, these avatars can also perform speech recognition, and sentiment analysis, powered by natural language processing (NLP); and have high accuracy and processing speed compared to those used in the physical world. Metaverse development and usage will result in big data analytics aspects, which will significantly increase the potential for data processing in the virtual world.

\section{Research Challenges and Perspectives}
In this section, inspired by the aforementioned architecture and key requirements of Metaverse, we discuss some challenges associated with the key technologies that Metaverse utilizes.

\subsection{
AI Models in Metaverse}
By applying AI, especially advanced ML algorithms, in the Metaverse, operators and designers can automate more efficiently and achieve higher performance than they could by using conventional approaches. Nevertheless, AI algorithms cannot be applied to facilitate users' operation or enhance the immersion experience. Currently available AI models are usually very deep and require large amounts of computation power, making them unsuitable for mobile devices which have limited resource availability. Therefore, it is necessary to design AI models that are lightweight yet efficient. In addition, DL algorithms are highly application-specific, so a model trained to solve one problem may not be able to solve another within a similar domain. To be used for other similar problems, the models must usually be retrained using the appropriate data. Additionally, bootstrapping is one of the most challenging aspects of the RL convergence process. Since most RL algorithms take a long time to converge, they are classified as NP-hard. It is also critical to note that time is of the essence when it comes to gaming, interactive experiences, and telemedicine in the Metaverse, and the system or user/avatar cannot accommodate delays. In order to design bootstrapping mechanisms that are efficient, more research is required. Further, looking forward, the next step of AI in the Metaverse should move from making hardware intelligent by AI algorithms to building intelligent hardware with AI algorithms, the latter being a hybrid mode and focusing on end-to-end solutions. Other advanced AI techniques, such as RL, federated learning, and few-shot learning, are likely to play a significant role in developing this solution.

\subsection{ 
Metaverse and Big Data}
As is well known, the Metaverse contains an enormous amount of data. Avatars, XR devices, or IoT devices may generate valuable but sensitive information from which meaningful solutions can be derived. In spite of this, serious issues regarding how data is captured, analyzed, and stored remain unresolved. The acquisition of data is the first step in data analysis, which is why it is crucial to collect accurate and realistic data with quality challenges. These data pose challenges in terms of quality. Besides data input from IoT devices, users and avatars also provide data. The Metaverse model will be misled if users produce low-quality content that is used for data analysis and model training. This is a large amount of data that must be stored in the Metaverse, and the amount of data is constantly increasing. Increasing data volumes present challenges with regard to storage capacity, expansion speed, and backup capability. The storage system, on the other hand, has to transform the model from a single file type to a variety of semi-structured and unstructured data relationships due to the diversity and complexity of user-generated content data and their interrelatedness. Considering the high infrastructure costs associated with these features, it is difficult for centralized storage methods to keep up with the volume and type of data. Several research and technological developments are needed to address this challenge, including communication, data interoperability, interactive experiences, storage, computing power, data sharing, and security and privacy.

\subsection{Metaverse for Mission Critical and Defense Applications}
Defense and aerospace industries are currently undertaking research to enhance defense and aerospace systems by developing a practical AR/MR/VR/XR Metaverse solution. A number of applications of the Metaverse in the defense and aerospace industries are effective for solving various problems, such as troubleshooting, inspecting aircraft and safety devices, and performing maintenance on aircraft. Moreover, by using Metaverse technology, real-world experiences can be automated and personalized to assist in the training of the Air Force and military personnel, as well as astronauts. It is also possible for AI-based Metaverses to enhance the security of both military and aerospace Metaverses. Therefore, the Metaverse plays a crucial role in training. The Metaverse, for instance, enhances the pilot and astronaut training process by providing many training scenarios and allowing pilots and astronauts to scale their ability to take swift action in difficult circumstances. Further, In Aerospace, virtual avatars are capable of providing real-time feedback on performance. However, Defense and aerospace pioneers must lead the next research axis in the Metaverse by adopting new strategies, analyzing their role in the digital world, identifying potential products and services, and establishing technical frameworks for improving key technologies necessary for the Metaverse.
\subsection{Metaverse and Efficiency of Security \& Privacy}
As a result of Metaverse's ability to collect data on behavior in a more detailed manner than user conversations and internet history, the confidentiality, integrity, availability, and authenticity of data remain under threat, and many serious unresolved issues remain. Due to this, security and privacy issues are of primary concern in the Metaverse. There is a need for avatar two-factor authentication and more protection of transmitted data, as well as a greater degree of vigilance in regard to a crime that may occur in the Metaverse. Wearable devices such as XR devices and IoT sensors play an essential role in creating an immersive experience. Additionally, their use opens up a greater number of opportunities for intrusions and attacks. Using a user's access history and stored information, a hacker can create a copy of the Metaverse account of a user. As outlined in a number of Metaverse literature related to security and privacy, different types of threats against the Metaverse and within it are presented, namely non-fungible token (NFT), Financial Fraud, AR/MR/VR/XR threats, darkverse, privacy issues, cyber-physical threats, social engineering threats, and traditional information and communication (IT) attacks. In order to meet the needs of future Metaverse applications, blockchain technologies, detection techniques, and countermeasures, AI requires extensive research in order to implement identity management, trust systems, and security monitoring systems. To enhance users' confidence in participating in the Metaverse and ensure financial and data security, it should concentrate particularly on preventing potential threat models against the Metaverse. Additionally, lightweight methods for ensuring data security must be developed in order to address the need for data security and privacy.
\subsection{Metaverse and Computing Power}
The concept of computing power is related to the ability to calculate, store, and transmit data. Computing power is one of the most important factors in developing a digital society, as it is a new factor of production. Developing the Metaverse environment requires the support of supercomputers. different technological requirements are integrated into the Metaverse, including AI, blockchain, and interaction technologies. Virtual and real worlds are also fused together. Thus, The Metaverse technology is highly reliant on IT technology, and it can not be accomplished without a powerful computing infrastructure. Therefore, there are still a number of challenges to be overcome in the development of the Metaverse, including both technological and energy challenges. As a consequence of the Metaverse's large-scale visualization of data and large-scale computing, computing power is required as well as electrical power. Furthermore, new solutions  are required to edge computing to extend cloud computing to the edge of networks and enable resource-limited devices to offload their tasks to edge servers so that they can be processed and then optimize their energy consumption.

\subsection{Metaverse Integration Across Disciplines} 
As a complex and dynamic system, the real world requires the cooperation of a variety of disciplines. Due to the fact that the Metaverse represents a virtual mapping of the real world, interdisciplinary research into the Metaverse is necessary. There are many aspects of modern life that are affected by the Metaverse, including medicine, defense, education, the economy, and so forth. In order to improve its own construction and to meet this challenge, relevant theoretical knowledge, research, and technologies must be developed. For instance, there is a strong correlation between the creation and maintenance of virtual currencies in the Metaverse and the real-world economic system, and their integration and development will have a profound impact on the real economy. Additionally, the Metaverse utilizes psychology and medicine based on advanced AI/ML algorithms to analyze and understand user behavior, which offers users a more immersive experience or can be used in education, training, psychotherapy, and other fields.
 
\section{Conclusion}
 This paper laid out a bold new vision for Metaverse that outlines the requirements, architecture, standards, status, challenges, and associated research. First, we provided the fundamental concepts of the Metaverse and discussed a variety of characteristics that allow the definition of several components of the Metaverse. Then, we discussed the existing virtual environment standards adopted in the Metaverse. We described Six technological requirements that have been provided for the Metaverse environment. For each key requirement, we have described the basic concepts and solutions that can be applied to address related issues. Furthermore, we discussed the current challenges and future perspectives of Metaverse, including AI applications, big data, Metaverse applications for mission-critical and defense applications, as well as security and privacy efficiency, computing power, and interdisciplinary aspects, and can therefore be referred to by researchers in future directions.

\balance


\vskip -0.9cm plus -1fil
\begin{IEEEbiographynophoto}	{Danda B. Rawat (S'07, M'09, SM'13)} is a Professor in the Department of Electrical Engineering \& Computer Science and Director of Research Institute for Tactical Autonomy (RITA) UARC, Director of DoD Centet of Excellence in AI/ML (CoE-AIML), Director of the Data Science \& Cybersecurity Center (DSC$^2$) at Howard University, Washington, DC, USA. Dr. Rawat is the recipient of NSF CAREER Award. 
\end{IEEEbiographynophoto}

\vskip -1cm plus -1fil
\begin{IEEEbiographynophoto}	{Hassan El Alami} is currently a Postdoctoral Fellow at Howard University, Washington DC, USA. He received his PhD in computer science and telecommunications from the National Institute of Posts and Telecommunications (INPT), Rabat, Morocco, in 2019. His current research interests include the Internet of Things, artificial intelligence and cybersecurity.

\end{IEEEbiographynophoto}

\end{document}